\documentclass{bmvc2k}

\title{IB-MVS: An Iterative Algorithm for Deep Multi-View Stereo based on Binary Decisions}

\addauthor{Christian Sormann}{christian.sormann@icg.tugraz.at}{1}
\addauthor{Mattia Rossi}{mattia.rossi@sony.com}{2}
\addauthor{Andreas Kuhn}{andreas.kuhn@sony.com}{2}
\addauthor{Friedrich Fraundorfer}{fraundorfer@icg.tugraz.at}{1}

\addinstitution{
Institute of Computer Graphics and Vision \\ Graz University of Technology \\ Austria
}
\addinstitution{
Sony Europe B.V. \\ R\&D Center - Stuttgart Laboratory 1 \\ Germany
}

\runninghead{Sormann, Rossi, Kuhn, Fraundorfer}{IB-MVS}

\usepackage{graphicx}
\usepackage{amsmath}
\usepackage{amssymb}
\usepackage{booktabs}
\usepackage{multirow}
\usepackage{placeins}
\usepackage[ruled,vlined]{algorithm2e}
\usepackage{capt-of}

\def\eg{\emph{e.g}\bmvaOneDot}

\begin{document} 

\maketitle

\begin{abstract}
We present a novel deep-learning-based method for Multi-View Stereo. Our method estimates high resolution and highly precise depth maps iteratively, by traversing the continuous space of feasible depth values at each pixel in a binary decision fashion. The decision process leverages a deep-network architecture: this computes a pixelwise binary mask that establishes whether each pixel actual depth is in front or behind its current iteration individual depth hypothesis. Moreover, in order to handle occluded regions, at each iteration the results from different source images are fused using pixelwise weights estimated by a second network. Thanks to the adopted binary decision strategy, which permits an efficient exploration of the depth space, our method can handle high resolution images without trading resolution and precision. This sets it apart from most alternative learning-based Multi-View Stereo methods, where the explicit discretization of the depth space requires the processing of large cost volumes. We compare our method with state-of-the-art Multi-View Stereo methods on the DTU, Tanks and Temples and the challenging ETH3D benchmarks and show competitive results.
\end{abstract}

\section{Introduction}

The objective of a Multi-View Stereo (MVS) system is the estimation of a dense depth map for a reference image, given one or multiple source images and all the camera poses.
This involves computing dense matching costs between the reference image and one or more source images.
In recent years, learning-based methods have shown promising results using learned input representations in the form of feature maps and learned similarity measures~\cite{mvsnet, casmvs,rmvsnet} for computing the matching costs.
However, most learning-based methods discretize the depth space and compute the matching cost at each selected depth for each reference image pixel.
The result is a cost volume whose size increases quadratically with respect to the image resolution for a given number of discretization steps.
As a result, cost volume methods are subject to computational and memory bottlenecks.
Newly proposed cascaded cost volume approaches~\cite{casmvs} mitigate these disadvantages, but they discretize the depth space using a pre-determined heuristic, which typically needs to be adapted for different datasets.

We propose a novel learning-based MVS method that explores the continuous depth space iteratively, without relying on an explicit cost-volume.
At each iteration, our method computes a pixelwise binary decision mask that estimates whether a given pixel actual depth is in front or behind its current depth hypothesis.
The binary decision permits to compute a new depth hypothesis at each pixel and the hypothesis is refined further at the next iteration.
Our work is inspired by the two-view stereo method in~\cite{bi3d_stereo}. However,~\cite{bi3d_stereo} assumes the same depth hypothesis for each pixel, estimating the depth by means of binary masks computed for a predefined set of depth values, thus building a cost volume.
Differently from~\cite{bi3d_stereo}, we do not construct a cost volume and rather propose a novel iterative architecture capable of estimating a binary decision mask for arbitrary depth hypotheses at individual pixels.
Moreover, our method targets the MVS scenario and takes advantage of the multiple source images available.
In particular, at each iteration, a binary decision mask is generated for each source image and the new resulting depth map hypotheses are fused with a learned weighting scheme inspired by~\cite{vismvsnet}. 
However, differently from~\cite{vismvsnet}, we do not fuse cost-volumes. Instead, we employ the weights to fuse 2D maps within an iterative depth estimation scheme.

We describe the proposed algorithm in detail in Section~\ref{sec_method}.
As our core contributions, we
1.) design a network architecture to estimate pixelwise depth dependent binary decision masks in the MVS setting,
2.) introduce a pixelwise depth inference algorithm
based on the prediction from the previous network,
3.)
implement a learning-based fusion strategy, inspired by ~\cite{vismvsnet}, for the depth maps predicted from different source images,
4.)
verify our results on the popular benchmarks DTU~\cite{dtu}, Tanks and Temples~\cite{tanksandtemples} and ETH3D~\cite{eth3d}.
\vspace{-1em}
\section{Related work}
In this section, we discuss the previously published related work in the MVS field and compare it with our proposed method. Traditional MVS methods rely on hand-crafted similarity measures such as normalized cross-correlation~\cite{colmap_mvs}.
The depth hypotheses space is typically explored via a plane-sweeping cost volume~\cite{plane_sweep_mvs} or the PatchMatch algorithm~\cite{colmap_mvs,gipuma,acmm, pcf_mvs}.
These methods also introduce techniques for pixelwise source view selection, in order to suppress the influence of matching results from occluded source images~\cite{colmap_mvs, acmm}.
The main limitation of these methods is represented by their hand-crafted similarity measures.
On the other hand, the PatchMatch exploration strategy for the depth hypothesis space and the employed pixelwise view-selection techniques elevate them above learning-based methods.

In recent years, deep-learning-based methods for MVS have received significant attention from the research community.
Earlier works focus on learning a feature representation and combine this with a learned similarity measure in the form of a 3D convolutional neural network~\cite{deepmvs,mvsnet}.
In order to reduce the computational cost of the regularization, recurrent~\cite{rmvsnet, dhcrmvsnet} and cascaded~\cite{casmvs, ucsnet} approaches were utilized.
The addition of CRF-based cost volume regularization was also explored in several works~\cite{bp_mvsnet,mvscrf}.
Other methods rely on a voxel-based representation of the input~\cite{surfacenet, surfacenet_plus} or refine an initial coarse estimate of a point cloud~\cite{pointmvsnet}.
The attention~\cite{attention_neurips} mechanism has been incorporated by several works~\cite{attmvsnet, lanet} as well.
More recent methods try to combine the benefits of traditional methods with the advantages of learned representation and similarity measures, \eg, by avoiding cost volumes and rather resorting to PatchMatch-based depth exploration strategies~\cite{patchmatchnet} or by leveraging pixelwise source view selection ~\cite{vismvsnet}. 

Similarly, our proposed method does not employ a cost volume for depth estimation, but still incorporates beneficial concepts from learning-based methods. 
In fact, it benefits from learned input representations and similarity measures, like learning-based approaches, and it implements an efficient strategy for the exploration of the depth hypothesis space, similarly to traditional methods.
Moreover, it employs a learning-based strategy to fuse estimates from different source images resembling the pixelwise view selection of traditional methods.
\vspace{-1em}
\section{Method} \label{sec_method}
In this section we present our MVS method, named \textit{IB-MVS} due its \textit{Iterative} approach and its relying on \textit{Binary} decisions.
Below, first we provide a system overview of IB-MVS, then we elaborate on the details of its novel depth inference algorithm and its network architectures.
\subsection{System overview}
\begin{figure}[t]
\centering
\includegraphics[width=1.0\textwidth]{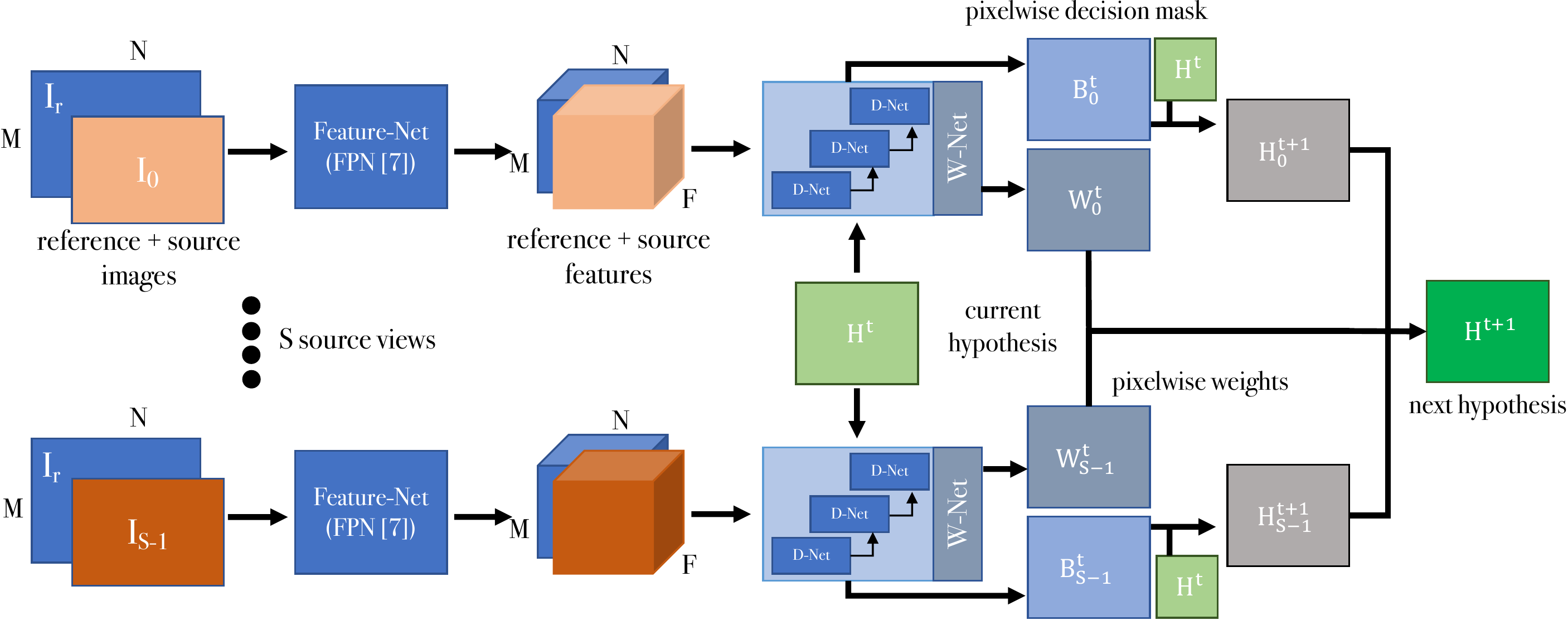}
\caption{Architecture of the overall system. Deep features extracted from the reference and source images using a FPN architecture~\cite{casmvs} are used as the input for three hierarchy levels of D-Nets, which predict the decision masks $B_s^t$. They are then used to compute the new inverse depth hypothesis $H_s^{t+1}$ for $s$. These are fused using weights $W_s^t$ estimated by W-Net and the fused result $H^{t+1}$ is the new inverse depth hypothesis for the next iteration.
\vspace{-10pt}
} 
\label{fig_arch_overview}
\end{figure}

The goal of our system is to estimate a dense depth map $d \in \mathbb{R}^{M \times N}$ of a reference image $I_r \in \mathbb{R}^{M \times N}$ given $S$ source images $I_s \in \mathbb{R}^{M \times N}$ with $s=0, 1, \ldots, S-1$.
Hereafter $(i, j)$ denotes a pixel location.
Our MVS method is iterative and traverses the continuous space of feasible depth values at each pixel in a binary decision fashion.
Inspired by~\cite{bi3d_stereo}, at each iteration $t$, and for each source image $I_s$, our method predicts a binary mask with the property:
\begin{equation} \label{eq_gt_mask}
    B_s^t(i,j)_{\text{GT}} = \begin{cases}
    1 & d_{\text{GT}}(i,j) < h^t(i,j), \\ 
    0 & \text{otherwise} 
    \end{cases}
\end{equation} 
where $h^t \in \mathbb{R}^{M \times N}$ is the current depth map hypothesis for the reference image and $d_{\text{GT}} \in \mathbb{R}^{M \times N}$ its ground truth depth map.
The pixelwise mask $B_s^t$ is predicted using a convolutional neural network named Decision Network (D-Net). 
In practice, we predict soft binary decision masks, hence $B_s^t(i, j)$ takes values in $[0, 1]$.
Each entry $B_s^t(i, j)$ permits to establish whether $d_{\text{GT}}(i, j)$ is in front or behind the current depth hypothesis $h^t(i, j)$.
Our method offsets the current depth hypothesis $h^t(i,j)$ and produces a new hypothesis $h_s^{t+1}(i,j)$ compliant with $B_s^t(i,j)$.
The $S$ new depth map hypothesis $h_s^{t+1}$, one for each source image, are then fused using learned pixelwise weights from a network named Weight Network (W-Net), in order to produce the next iteration depth map hypothesis $h^{t+1}$.
We depict IB-MVS in Figure~\ref{fig_algo_vis} and provide a visual overview of the entire system in Figure~\ref{fig_arch_overview}.
\subsection{Depth inference algorithm} \label{sec_method_inference}
The depth inference algorithm of IB-MVS assumes a feasible depth range $[d_{\text{min}}, d_{\text{max}}] \in \mathbb{R}^2$ as input and computes the reference image depth map $d \in \mathbb{R}^{M \times N}$ iteratively.
The algorithm operates in the inverse depth domain, as this yields improved results in scenes with large depth ranges.
To this end, we introduce the inverse depth map hypothesis $H = 1 / h \in \mathbb{R}^{M \times N}$ and define the inverse depth range bounds $D_{\text{min}} = 1 / d_{\text{min}}$ and $D_{\text{max}} = 1 / d_{\text{max}}$.

\begin{figure}[t]
\centering
\includegraphics[width=0.99\textwidth]{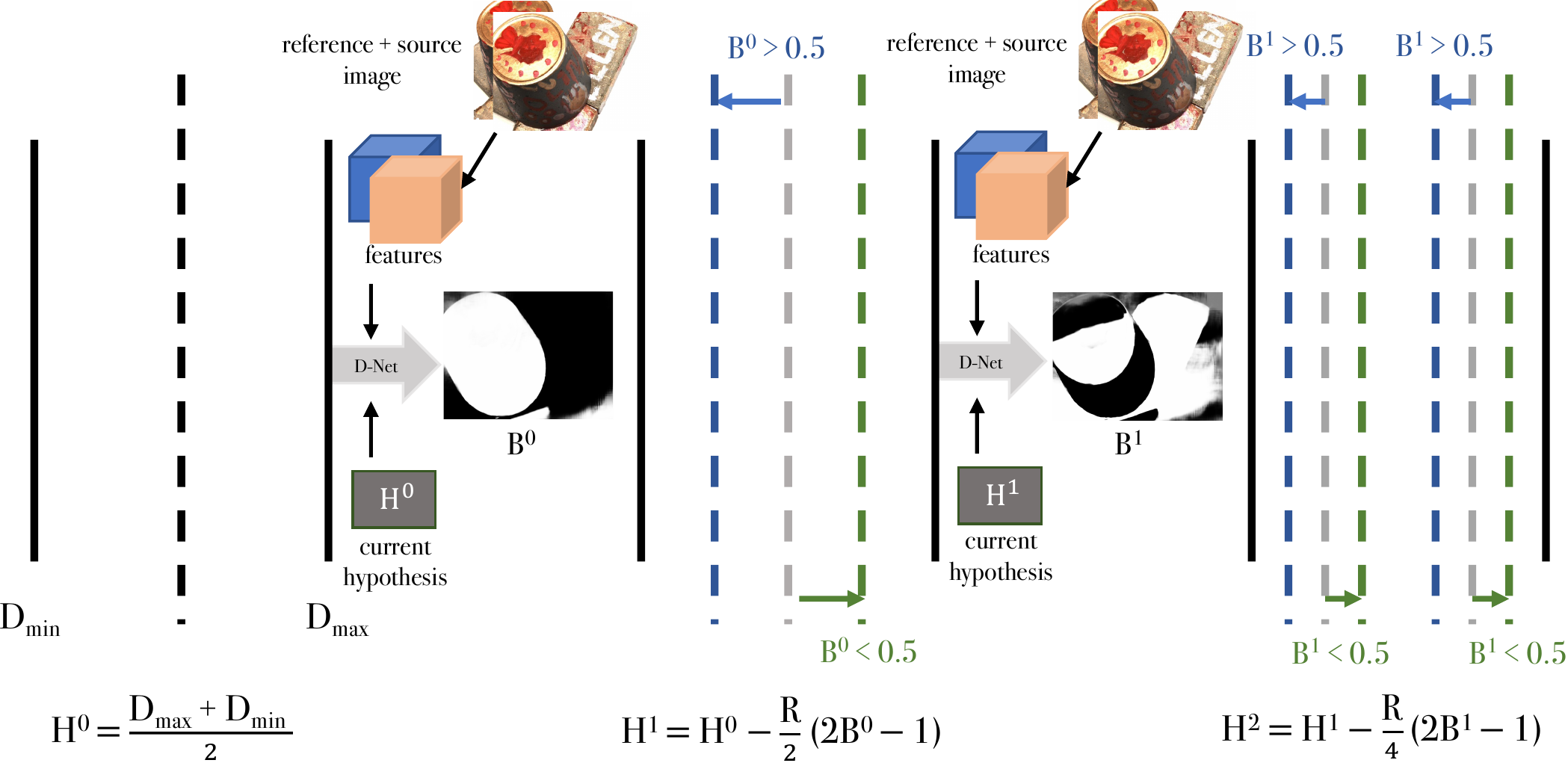}
\caption{
Visualization of IB-MVS depth inference algorithm for a single source image.
The inverse depth range $[D_{\text{min}}$, $D_{\text{max}}]$ is explored iteratively by updating the inverse depth map hypothesis $H^t$ by means of the binary decision mask $B^t$ with $t = 0, 1, \ldots, T-1$.
\vspace{-10pt}
}
\label{fig_algo_vis}
\end{figure}

The algorithm starts at iteration $t=0$ by setting $H^0(i,j) = \frac{D_{\text{max}} + D_{\text{min}}}{2}$ at each pixel.
It then uses D-Net with $H^0$ as inverse depth hypothesis in order to compute the first set of binary decision masks $B_s^0(i,j)$, one for each source image.
The next pixelwise depth map hypotheses at iteration $t+1$ is then calculated as follows:
\begin{equation} \label{eq_hypo_update}
   H_s^{t+1}(i,j) = H^t(i,j) - \frac{R}{2^{t+1}} (2 B_s^t(i,j) - 1), \quad s = 0, 1, \ldots, S-1
\end{equation}
where $R = \frac{D_{\text{max}} - D_{\text{min}}}{2}$ and $\Delta R^{t}=\frac{R}{2^{t+1}}$ is referred as the \textit{step size}.
The update in Eq.~\eqref{eq_hypo_update} is guided by the binary decision mask $B_s^t$, estimated from the current inverse depth hypothesis $H^t$.
For $B_s^t(i,j)=1$, the sought ground truth depth is in front of the current hypothesis, therefore we step backwards.
For $B_s^t(i,j)=0$, we step forward instead.
Since $B_s^t(i, j)$ takes values in $[0, 1]$, the step size performs a smooth update of the current depth hypothesis.
The step size $\Delta R^{t}=\frac{R}{2^{t+1}}$ decreases at each iteration, which represents the halving of the search space.
However, it is noteworthy that the magnitude of the hypothesis update is adaptive (in both directions) thanks to its dependence on $B_s^t(i,j)$, as modeled in Eq.~\eqref{eq_hypo_update}.
The update is sketched in Figure~\ref{fig_algo_vis}, where for the sake of simplicity one source image is assumed, hence omitting subscript $s$.
After computing the new inverse depth map hypothesis $H_s^{t+1}(i,j)$ for each source image $s$, the fused inverse depth map hypothesis $H^{t+1}\in \mathbb{R}^{M \times N}$ is calculated as:
\begin{equation} \label{eq_weight_fusion}
   H^{t+1}(i,j) = \frac{1}{W^t(i,j)} \sum_{s=0}^{S-1} W_s^t(i,j) H_s^{t+1}(i,j)
\end{equation} 
where $W^t(i,j) = \sum_{s=0}^{S-1} W_s^t(i,j)$ is the sum of the weights estimated by W-Net for each source image. In~\cite{vismvsnet} this normalization was argued to be more beneficial than thresholding.
We perform $T$ iterations and set the final depth map estimate $d = h^{T} = 1 / H^{T}$.

\subsection{Binary decision network}

The decision network D-Net exhibits a U-Net~\cite{unet}-like encoder-decoder structure and it is depicted in Figure~\ref{fig_dnet_overview}.
At the top level, the decoder is fed both with the feature maps $F_r$ and $F_s \in \mathbb{R}^{F \times M \times N}$, obtained by applying a Feature Pyramid Network~\cite{casmvs} to the reference and source images $I_r$ and $I_s$, respectively, as well as with a depth map hypothesis $h \in \mathbb{R}^{M \times N}$.
At each level of the encoder, deformable convolutions~\cite{deform_convsv2} are used to convolve the source image feature maps along locations on the epipolar line determined by the depth hypothesis.
Specifically, we deform a $k \times k$ kernel such that each of the $k^2$ sampling locations of the kernel correspond to locations on the epipolar line.
We use kernel size $k=5$ and can thus center the deformed kernel on the epipolar line at the location predicted by the current depth hypothesis $h^t(i,j)$ and distribute the samples on either side spaced with a unit vector in pixel coordinates.
The resulting sampled features from the source feature map are then concatenated with the reference feature map, as suggested in~\cite{bi3d_stereo}.
This procedure is repeated at each resolution level, the resulting feature maps are further processed with standard convolutional layers and passed both to the next lower resolution level of the encoder and to the decoder, as depicted in Figure~\ref{fig_dnet_overview}.
At the decoder side, the feature maps are upsampled and further concatenated with the features from the encoder at the next higher resolution level.
The output binary mask $B^t_s$ is generated using a sigmoid activation function. 
We include a detailed specification of the convolutional hyper-parameters in Section A of the supplementary material.
In practice, in order to predict the binary decision mask $B_s^t$, we employ three D-Nets on full, half and quarter resolution inputs.
In particular, as depicted in Figure~\ref{fig_dnet_overview}, the output features from the previous scale D-Net are employed on the next scale. 
We observed that employing three resolution levels leads to higher quality binary decision masks. In fact, this choice increases the overall architecture receptive field and permits a coarse to fine refinement of the estimated masks, as each level employs the previous level output.
\subsection{Fusion weights network}
\begin{figure}[t]
\centering
\includegraphics[width=0.99\textwidth]{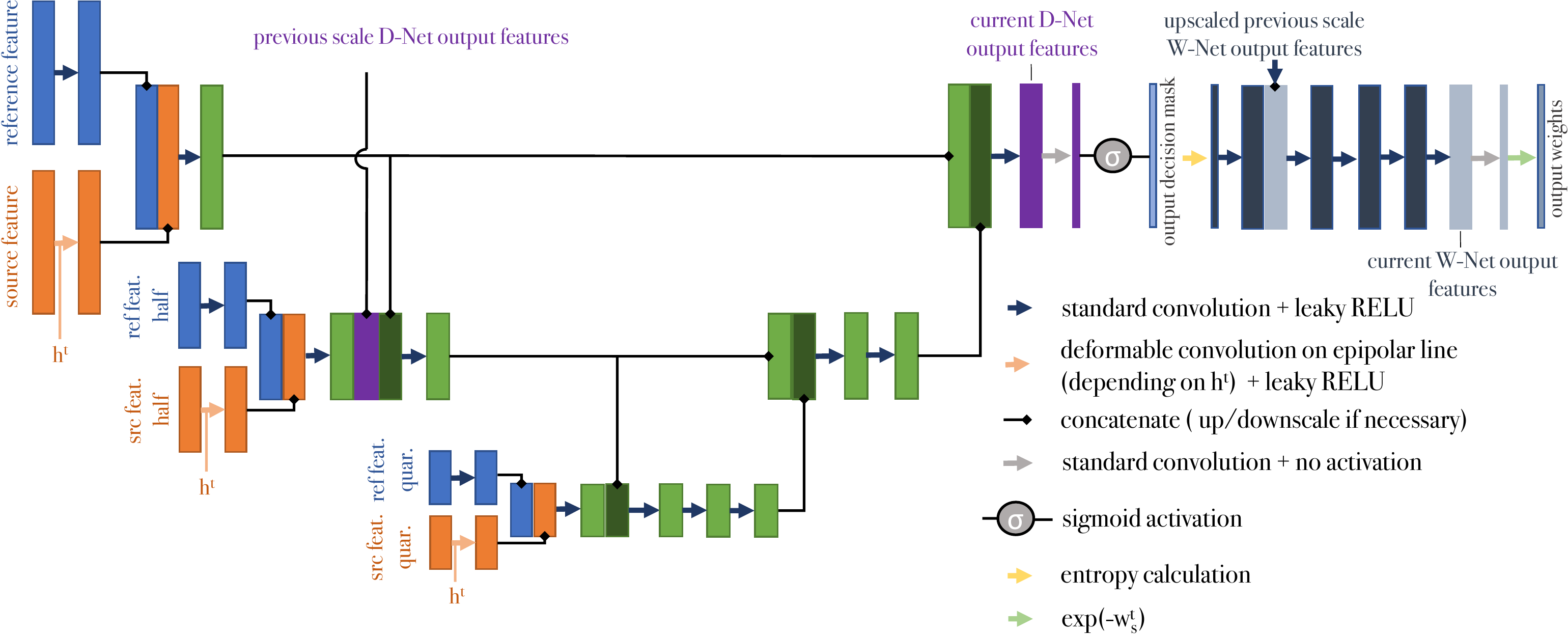}
\caption{
D-Net and W-Net architecture overview showing feature channel slices.
The network uses standard 2D convolutions in combination with deformable 2D convolutions~\cite{deform_convsv2} which utilize depth hypothesis $h^t$ to sample the epipolar line of the source image.
\vspace{-10pt}
}
\label{fig_dnet_overview}
\end{figure}

In the proposed architecture, each D-Net is followed by a W-Net, a network whose objective is to assign a confidence, in the form of a weight map, to the predicted binary decision mask.
Figure~\ref{fig_dnet_overview} depicts W-Net in gray.
Inspired by~\cite{vismvsnet}, W-Net operates on the pixelwise entropy of the predicted binary mask $E_s^t = -(B_s^t \log(B_s^t) + (1 - B_s^t) \log(1 - B_s^t))$.
This design strategy, coupled with the choice to use the depth hypothesis to sample the source image, rather than as a network input, makes our overall algorithm scale independent.
As suggested in~\cite{vismvsnet}, the final weight at the pixel $(i, j)$ is predicted as $W_s^t(i,j) = \exp(-w_s^t(i,j))$ where $w_s^t(i,j)$ is the network output.
The pixelwise weights are used by our depth inference algorithm during the fusion stage in Eq.~\eqref{eq_weight_fusion}.
The ideally predicted weights are small in those areas where the binary decision mask is not reliable, such as in occluded regions, and large otherwise, such that inverse depth hypotheses from different source images can complement each other.
This approach mitigates the negative effect of potentially erroneous estimates $H_s^{t+1}$ from the source views, when fusing them into the new depth hypothesis $H^{t+1}$.
This is crucial, as $H^{t+1}$ represents the starting point for the estimation of the next iteration binary decision masks. 
\vspace{-1em}
\section{Experiments}
In this section, firstly we describe the training procedure adopted for IB-MVS, then we compare it to state-of-the-art works on popular MVS benchmarks.
\subsection{Network training} \label{sec_network_train}
First, the multi-level D-Net is pre-trained on random uniform inverse depth map hypotheses with $H(i,j) = D_{\text{rand}}$ for every $(i, j)$ and $D_{\text{rand}} \in [D_{\text{min}},D_{\text{max}}]$, as proposed in~\cite{bi3d_stereo}.
The ground truth binary decision mask $B_{\text{GT}} \in \mathbb{R}^{M \times N}$ for a given depth map hypothesis $h = 1 / H$ is computed from the ground truth depth map $d_{\text{GT}}$ using Equation~\eqref{eq_gt_mask}.
We employ a loss $L_k$ at each level $k = 0, 1, 2$ of the multi-level D-Net, with $2$ being the full resolution.
In particular, $L_k$ is defined as the average of the Binary Cross Entropy (BCE) at the valid pixels $(i, j)$:
\vspace{-10pt}
\begin{equation} \label{eq_loss}
    L_k = \frac{1}{\overline{V}} \sum_{i,j} V(i,j) \> \text{BCE}(B(i,j), B_{\text{GT}}(i,j))
    \vspace{-5pt}
\end{equation} 
where $V = \sum_{i, j} V(i, j)$ with $V(i, j)$ equal to $1$ if the pixel has a valid ground truth depth, $0$ otherwise.
The BCE loss in Eq.~\eqref{eq_loss} is defined as follows:
\begin{equation}
    \text{BCE}(b, b_{\text{GT}}) = -(b_{\text{GT}} \log(b) + (1 - b_{\text{GT}}) \log(1 - b)).
\end{equation}
The overall loss is the weighted sum $L = \sum_{k=0}^{2} \lambda_k L_k$ with $\lambda_{0,1,2} = (0.25, 0.5, 1.0)$.
After the pre-training, we train the multi-level D-Net using the depth inference algorithm described in Section~\ref{sec_method_inference} with $T=8$ iterations.
However, we do not consider the fusion step at this stage and work with a single randomly selected source image over the $T$ iterations. 
The random selection of source images was shown to be beneficial by~\cite{patchmatchnet}.
We employ the previously introduced loss $L$ on each one of the binary decision masks $B_s^t$, with $t=0, 1, ..., T-1$, generated alongside the depth inference procedure.
We refer to the loss at the iteration $t$ as $L^t$.
At iteration $t$, the inverse depth hypothesis $H^t$ is used both by the multi-level D-Net to generate $B_s^t$ and in Eq.~\eqref{eq_gt_mask} to generate its ground truth.
Now the loss $L^t$ can be computed and the next hypothesis $H^{t+1}$ generated.
The procedure is iterated and the final loss is the sum over the $T$ losses $L^t$.
We do not back-propagate across iterations. 
Finally, in the third training stage, we train W-Net jointly with D-Net.
At this stage we fuse $B_s^t$ from $4$ randomly selected source images using the weights from W-Net with the same approach used to fuse the inverse hypotheses in Section~\ref{sec_method_inference} to generate $B^t$.
We compute the loss $L^t$ of one iteration, where $t$ is randomly chosen in $\{0, \dots, T-1\}$, as the sum of losses on the $S$ individual $B_s^t$ and fused $B^t$. 
We implemented IB-MVS in PyTorch~\cite{pytorch} and trained with batch size $1$ using ADAM~\cite{adam_optimizer}. 
\subsection{Evaluation metrics}
In our evaluation, we present experimental results on three popular MVS benchmarks, namely DTU~\cite{dtu}, Tanks and Temples~\cite{tanksandtemples} and ETH3D high and low-res~\cite{eth3d}. These benchmarks compare the reconstructed point cloud against a dense ground truth and extract completeness and accuracy metrics (recall and precision, respectively, for Tanks and Temples~\cite{tanksandtemples}). Completeness and accuracy are aggregated into a single metric: their average for DTU~\cite{dtu} and harmonic mean, denoted F-score, for Tanks and Temples~\cite{tanksandtemples} and ETH3D~\cite{eth3d}. For DTU~\cite{dtu}, accuracy and completeness are measured in mm, hence lower is better. For ETH3D~\cite{eth3d} and Tanks and Temples~\cite{tanksandtemples}, these metrics are percentages, hence higher is better.
\subsection{Ablation study}
We first investigate the influence of the iterations $T$ on the point cloud results for DTU~\cite{dtu}. 
In Table~\ref{tab_ablation}, we show that increasing the number of iterations leads to improved results, which is coherent with our iterative depth refinement. 
The larger performance difference between $6-7$ iterations, compared to $7-8$ and $8-9$, is explained by the progressively shrinking step size of IB-MVS.
Furthermore, we investigate the benefits of using our W-Net in the fusion step. 
In particular, we compare it to a naive strategy that simply averages the depth map hypotheses from the different source views. 
The results in Table~\ref{tab_ablation} show that, for the same number of iterations $T = 8$, W-Net leads to a better completeness and an overall better quality (avg. metric) than the naive fusion strategy, while exhibiting a competitive accuracy.
The improved accuracy of the naive fusion strategy is obtained at the cost of worse completeness, as the absence of W-Net leads to more inconsistent estimates in occluded regions.
\begin{table}[t] 
\begin{minipage}{0.43\textwidth}
  \centering
  \setlength{\tabcolsep}{2pt}
  \small
  \begin{tabular}{@{}cccccc@{}}
      \toprule
      W-Net & T & avg. & acc. & cmp. & RT \\ \midrule
       \checkmark & 6 & 0.717 & 0.769 & 0.664 & 2.0s  \\
       \checkmark  & 7 & 0.371 & 0.402 & 0.340 & 2.3s  \\
       \checkmark  & 8 & \textbf{0.321} & 0.334 & \textbf{0.309} & 2.7s  \\ \hline
        - & 8 & 0.342 & 0.326 & 0.359 & 2.3s  \\ 
        - & 9 & 0.343 & \textbf{0.324} & 0.362 & 2.6s \\
      \bottomrule
      \end{tabular}
  \caption{Ablation study on DTU test-set using accuracy, completeness and their average in mm (lower is better). We observe improvements with increasing iterations T. After T=8 the subdivision of the search space is sufficient, thus the result for T=9 is very close. Further, the inclusion of W-Net improves the results for the same number of iterations.}
  \label{tab_ablation}
  \end{minipage} \hfill
  \begin{minipage}{0.55\textwidth}
      \centering
      \includegraphics[width=1.0\textwidth]{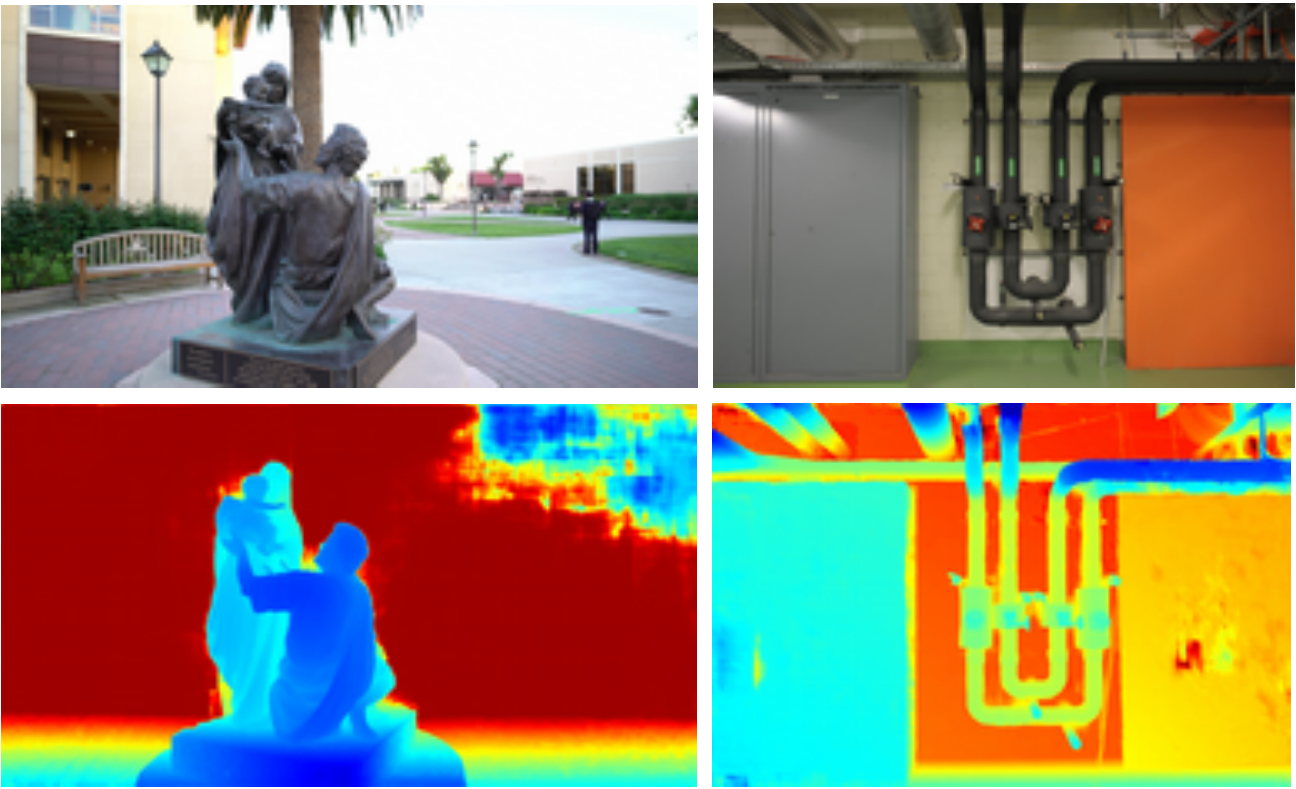}
      \captionof{figure}{Qualitative depth map results from Tanks and Temples~\cite{tanksandtemples} and ETH3D~\cite{eth3d}.\vspace{-10pt}}
      \label{fig_depth_map_res}
  \end{minipage}
  \end{table}
\subsection{Benchmark results}
We use Tanks and Temples~\cite{tanksandtemples} and ETH3D~\cite{eth3d} high and low-res in order to evaluate the generalization capabilities of IB-MVS, since we do not train on their respective training sets.

The feasible depth range $[d_{min}, d_{max}]$ is inferred from the SfM model using the method in~\cite{mvsnet}.
The computed depth maps are fused into a single point cloud with the proposed method of~\cite{mvsnet}; we denote its parameters representing the number of consistent views and the geometric re-projection error threshold as $S_g$ and $g$, respectively. 
We denote the used image resolution as $M \times N$, the runtime per image as $RT$ and the memory consumption as $MEM$.
Finally, we recall that $S$ is number of used source views.

For the evaluation on the DTU benchmark, we train the network on DTU for ($8$, $12$, $2$) epochs and learning rates ($10^{-4}$, $10^{-4}$, $10^{-5}$) in the stages ($1$ $2$, $3$) described in Section~\ref{sec_network_train}, respectively.
For the evaluations on Tanks and Temples and ETH3D, we train on DTU at stage $1$ and on Blended MVS~\cite{blended_mvs} at stages $2$ and $3$, for ($8$, $21$, $4$) epochs using learning rates ($10^{-4}$, $10^{-4}$, $10^{-5}$).
Every training on DTU employs the ground truth depth maps and train-test split of~\cite{mvsnet}.
Finally, our method is run for $T=8$ iterations on DTU and for $T=9$ iterations on Tanks and Temples~\cite{tanksandtemples} and ETH3D~\cite{eth3d}. 
All the experiments were performed using an Nvidia RTX 2080Ti graphics card.
\begin{figure}[t]
\centering
\includegraphics[width=1.0\textwidth]{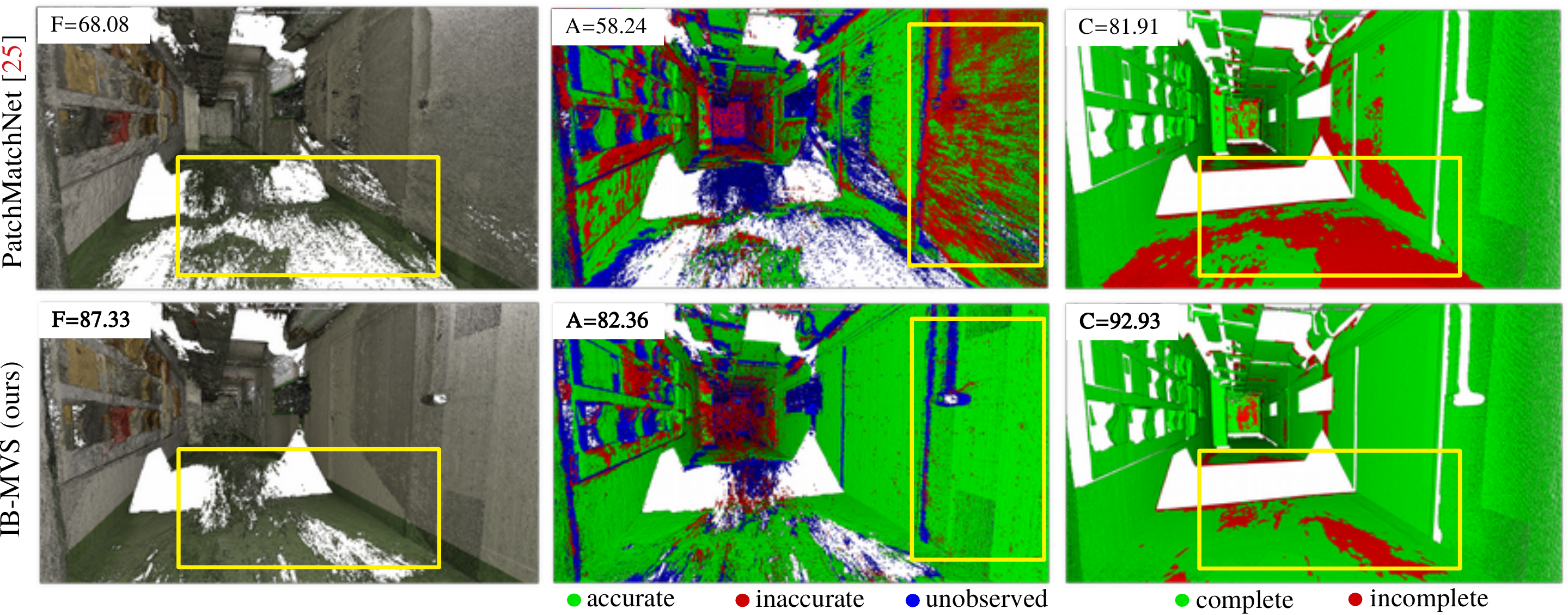}
\vspace{-20pt}
\caption{
Point cloud comparison with PatchMatchNet~\cite{patchmatchnet} on the ETH3D~\cite{eth3d} high-res terrains dataset. We show the output point cloud, accuracy and completeness.
\vspace{-8pt}
}
\label{fig_point_cloud_results}
\end{figure}
\begin{table}[t]
\resizebox{\textwidth}{!}{
\centering
\small
\setlength{\tabcolsep}{1.5pt}
\begin{tabular}{@{}lcccccccccccccc@{}}
 \toprule
 & \multicolumn{4}{c}{low-res-test} & \multicolumn{3}{c}{low-res-train}  & \multicolumn{4}{c}{high-res-test} & \multicolumn{3}{c}{high-res-train}  \\
 \cmidrule{2-15} 
 publication & F & acc. & cmp. & RT & F & acc. & cmp. & F & acc. & cmp. & RT & F & acc. & cmp. \\ \midrule
 DeepC-MVS~\cite{deepcmvs} & \underline{62.37} & \underline{65.89} & 59.42 & 5746 & \underline{61.99} & 65.98 & \underline{59.27} & \underline{87.08} & 89.15 & \underline{85.52} & 3155 & \underline{84.81} & 90.37 & \underline{80.30}   \\  
 COLMAP~\cite{colmap_mvs} & 52.32 & 61.51 & 45.89 & 3312 & 49.91 & \underline{69.58} & 40.86 & 73.01 & \underline{91.97} & 62.98 & 1658 & 67.66 & \underline{91.85} & 55.13   \\  
 ACMM~\cite{acmm} & 55.01 & 52.37 & 58.27 & 1662 & 55.12 & 54.69 & 57.01 & 80.78 & 90.65 & 74.34 & 1165 & 78.86 & 90.67 & 70.42   \\  
 PCF-MVS~\cite{pcf_mvs} & 57.06 & 56.56 & 58.42 & 9289 & 57.32 & 57.03 & 58.17 & 80.38 & 82.15 & 79.29 & 2272 & 79.42 & 84.11 & 75.73   \\  \hline
 R-MVSNet~\cite{rmvsnet} & 36.87 & 37.45 & 37.16 & 2413 & - & - & - & - & - & - & - & - & - & -  \\  
 CasMVSNet~\cite{casmvs} & 44.49 & 55.44 & 38.80 & - & 49.00 & \textbf{62.06} & 41.86 & - & - & - & - & - & - & -   \\ 
 MVSCRF~\cite{mvscrf} & 28.32 & 34.84 & 24.97 & - & - & - & - & - & - & - & - & - & - & -   \\
 P-MVSNet~\cite{pmvsnet} & 44.46 & 54.95 & 38.28 & \underline{627} & - & - & - & - & - & - & - & - & - & -    \\  
 BP-MVSNet~\cite{bp_mvsnet} & 43.22 & 32.65 & 64.34 & - & 50.87 & 49.12 & \textbf{55.29} & - & - & - & - & - & - & -  \\ 
 Att-MVS~\cite{attmvsnet} & 45.85 & \textbf{64.84} & 37.07 & - & - & - & - & - & - & - & - & - & - & -  \\  
 PVSNet~\cite{pvsnet} & 45.78 & 38.39 & 57.76 & 2116 & - & - & - & 72.08 & 66.41 & 80.05 & 830 & 67.48 & - & - \\
 PMNet~\cite{patchmatchnet} & - & - & - & - & - & - & - & 73.12 & 69.71 & 77.46 & \underline{493} & 64.21 & 64.81 & 65.43  \\ 
 IB-MVS (ours) & \textbf{49.19} & 39.31 & \underline{67.29} & 1487 & \textbf{55.84} & 61.06 & 52.66 & \textbf{75.85} & \textbf{71.64} & \textbf{82.18} & 616 & \textbf{71.21} & \textbf{75.21} & \textbf{69.02} \\
 \bottomrule
 \end{tabular}}
\caption{
ETH3D~\cite{eth3d} results: F-score, accuracy and completeness as percentage (2cm).
Higher is better, overall best underlined, best among learning based in bold. For the test sets, we report the average per-scene runtime RT in seconds (for methods that provide this).
\vspace{-15pt}
}
\label{tab_eth_results}
\end{table}
\begin{table}[t]
    \centering
    \setlength{\tabcolsep}{3pt}
    \begin{tabular}{@{}lccccccccc@{}}
    \toprule
    & \multicolumn{3}{c}{\textbf{DTU} \cite{dtu}} & \multicolumn{6}{c}{\textbf{Tanks and Temples} \cite{tanksandtemples}} \\ 
    & \multicolumn{3}{c}{test} & \multicolumn{3}{c}{intermediate} & \multicolumn{3}{c}{advanced}\\ 
    \cmidrule{2-10} 
    publication & avg. & acc. & \multicolumn{1}{c}{cmp.} & F & preci. & \multicolumn{1}{c}{reca.} & F & preci. & reca.  \\ \midrule
    DeepC-MVS~\cite{deepcmvs} & - & - & - & 59.79 & 59.11 & 61.21 & 34.54 & \underline{40.68} & 31.30 \\  
    ACMM~\cite{acmm} & - & - & - & 57.27 & 49.19 & 70.85 & 34.02 & 35.63 & 34.90 \\  
    COLMAP~\cite{colmap_mvs} & - & - & - & 42.14 & 43.16 & 44.48 & 27.24 & 33.65 & 23.96    \\  
    PCF-MVS~\cite{pcf_mvs} & - & - & - & 55.88 & 49.82 & 65.68 & \underline{35.69} & 37.52 & 35.36  \\  
    \hline
    R-MVSNet~\cite{rmvsnet} & 0.422 & 0.385 & 0.459 & 48.40 & 43.74 & 57.60 & 24.91 & 31.47 & 22.05 \\  
    BP-MVSNet~\cite{bp_mvsnet} & 0.327 & 0.333 & 0.320 & 57.60 & 51.26 & 68.77 & 31.35 & 29.62 & 35.61 \\ 
    P-MVSNet~\cite{pmvsnet} & 0.420 & 0.406 & 0.434 & 55.62 & 49.93 & 63.82 & - & - & -   \\  
    Att-MVS~\cite{attmvsnet} & 0.356 & 0.383 & 0.329 & \underline{60.05} & \underline{61.89} & 58.93 & 31.93 & \textbf{40.58} & 27.26  \\  
    CVP-MVSNet~\cite{cvp_mvsnet} & 0.351 & \textbf{0.296} & 0.406 & 54.03 & 51.41 & 60.19 & - & - & -  \\
    CasMVSNet~\cite{casmvs} & 0.355 & 0.325 & 0.385 & 56.84 & 47.62 & 74.01 & 31.12 & 29.68 & 35.24  \\  
    PatchMatchNet~\cite{patchmatchnet} & 0.352 & 0.427 & \textbf{0.277} & 53.15 & 43.64 & 69.37 & \textbf{32.31} & 27.27 & \underline{41.66} \\
    UCS-Net~\cite{ucsnet} & 0.344 & 0.338 & 0.349 & 54.83 & 46.66 & 70.34 & - & - & - \\
    LANet~\cite{lanet} & 0.335 & 0.320 & 0.349 & 55.70 & 45.62 & \underline{75.68} & - & - & - \\
    D2HC-RMVSNet~\cite{dhcrmvsnet} & 0.386 & 0.395 & 0.378 & 59.20 & 49.88 & 74.08 & - & - & - \\
    VisMVSNet~\cite{vismvsnet} & 0.365 & 0.369 & 0.361 & 60.03 & 54.44 & 70.48 & - & - & - \\ 
    IB-MVS (ours) & \textbf{0.321} & 0.334 & 0.309 & 56.02 & 47.71 & 72.64 & 31.96 & 27.85 & 41.48 \\
    \bottomrule
    \end{tabular}
    \caption{
    DTU~\cite{dtu} results: accuracy, completeness and their average are in mm, lower is better.
    Tanks and Temples~\cite{tanksandtemples} results: precision, recall and F-score are percentages, higher is better.
    Overall best results are underlined, best among learning based methods are bold.
    \vspace{-10pt}
    }
    \label{tab_dtu_tanks_results}
\end{table}
\paragraph{ETH3D benchmark~\cite{eth3d}}
This dataset is the most challenging one for learning-based MVS methods, especially the high-res subset.
On the one hand, the high resolution of its images represents a memory bottleneck for learning-based cost-volume methods.
On the other, it is characterized by images with wide baselines and with a significantly lower overlap than in Tanks and Temples~\cite{tanksandtemples} and DTU~\cite{dtu}, which can make matching without pixelwise source view selection difficult.
For the high-res dataset we set $M \times N=1984 \times 1312$, $S=4$, $S_g=1$, $g=1.0$ and obtain $RT=7.7s$, $MEM=7.9GB$.
For the low-res dataset we set $M \times N=928 \times 512$, $S=4$, $S_g=3$, $g=0.1$ and obtain $RT=1.5s$, $MEM=2.3GB$.
We provide quantitative results for both the datasets in Table~\ref{tab_eth_results}.
Although this benchmark had been dominated by traditional methods such as ACMM~\cite{acmm} and DeepC-MVS~\cite{deepcmvs} in the past, the recently published PatchMatchNet~\cite{patchmatchnet} was able to achieve competitive results: IB-MVS outperforms~\cite{patchmatchnet} on both the training and the test datasets. 
Learning-based approaches relying on cost-volumes are limited on this benchmark, however our iterative approach allows IB-MVS to infer accurate results, even on high resolution images. Further, since large viewpoint changes are present in this benchmark, the implemented fusion scheme allows IB-MVS to deal with occluded regions.
\paragraph{DTU benchmark~\cite{dtu}}
This dataset contains close-range images of various objects.
For this dataset, we set $M \times N=1152 \times 864$, $S=4$, $S_g=3$, $g=0.25$ and obtain $RT = 2.7s$, $MEM = 3.8GB$.
In Table~\ref{tab_dtu_tanks_results} we compare IB-MVS to the recent learning-based state-of-the-art methods.
It can be observed that IB-MVS provides very competitive accuracy and completeness values: in particular it achieves a good trade-off between the two, which results in the best average score.
Finally, we achieve a complete reconstruction, even with a strict filtering parameter $g$.
This is due to IB-MVS ability to yield very precise results without the need to use large cost-volumes. This is especially important in the case of the DTU benchmark, where we target highly precise reconstructions of single objects.
\paragraph{Tanks and Temples benchmark~\cite{tanksandtemples}}
The intermediate subset focuses on reconstructing small and large single objects, while the advanced subset consists of large scale indoor and outdoor scenes.
In our experiments, we set $M \times N=1920 \times 1056$, $S=4$, $S_g=4$, $g=0.5$ (interm.), $S_g=3$, $g=0.5$ (adv.) and obtain $RT=6s$, $MEM=6.4GB$.
On the advanced subset, IB-MVS is competitive with other state-of-the-art learning-based methods such as CasMVSNet~\cite{casmvs} and PatchMatchNet~\cite{patchmatchnet}, as shown in Table~\ref{tab_dtu_tanks_results}.
Our iterative exploration of the hypothesis space, along with our pixelwise source view fusion, allows IB-MVS to achieve competitive results among learning based methods on the challenging advanced subset.
A possible direction for improving the intermediate set results could be to employ a confidence measure for the final predicted depth map, to filter out inaccurate points ahead of the point cloud fusion, similarly to~\cite{deepcmvs}.
Another direction could consider the improvement of the core architecture, for instance, by incorporating a regularization stage. In particular, the binary decision mask prediction could be regularized by a differentiable CRF, such as~\cite{bp_mvsnet}.

Finally, we discuss IB-MVS runtime performance on ETH3D, measured as the per-scene runtime of the complete reconstruction, in seconds, including the point cloud fusion step.
IB-MVS is faster than traditional methods on both ETH3D high-res and low-res, as shown in Table~\ref{tab_eth_results}. It is noteworthy that the low-res scenes contain more images.
While the learning-based method PM-Net~\cite{patchmatchnet} yields a better runtime than IB-MVS, a direct comparison is difficult. In fact, the PatchMatch algorithm utilized within PM-Net~\cite{patchmatchnet} operates at half-resolution and the full resolution is obtained via a subsequent up-sampling and refinement.
Instead, IB-MVS operates at full resolution and does not perform an additional refinement. Furthermore, IB-MVS achieves a better F-score than PM-Net~\cite{patchmatchnet} on ETH3D high-res.
On low-res, IB-MVS offers a competitive runtime compared to learning-based methods. In terms of F-score, IB-MVS outperforms the fastest learning-based method P-MVSNet~\cite{pmvsnet}, which does not participate on high-res.
\vspace{-10pt}
\section{Conclusion}
We presented IB-MVS, a novel learning-based approach for MVS that explores the depth space iteratively in a binary decision fashion.
IB-MVS couples the advantages of learning-based methods, such as learned input representations, with an efficient exploration strategy of the hypothesis space.
In fact, IB-MVS can handle high resolution images, as it does not require a cost volume.
In addition, IB-MVS benefits from a pixelwise source view fusion strategy.
Extensive results show that IB-MVS achieves competitive results compared to state of the art methods on popular MVS benchmarks. \textbf{Acknowledgement}: This work has been supported by the FFG, Contract No. 881844: "Pro$^2$Future". 

\FloatBarrier

\setcounter{secnumdepth}{0}

\section{Supplementary material}

\renewcommand\thesection{\Alph{section}}
\setcounter{secnumdepth}{10}
\setcounter{section}{0}

\section{Network architecture hyper-parameters}

We provide the network hyper-parameters of D-Net and W-Net in Table~\ref{tab_network_arch_dnet} and Table~\ref{tab_network_arch_wnet}, respectively.
As specified in the main paper, we use three resolution levels $l=(0,1,2)$ at quarter, half and full resolution. For generating the image features $\text{Feat}_{r_l}$ and $\text{Feat}_{s_l}$, we utilize the FPN architecture of~\cite{casmvs} and set the number of feature channels for each level $F_l = (32, 16, 8)$. Further, we replace the batch normalization~\cite{batch_norm} in the FPN~\cite{casmvs} with instance normalization~\cite{instance_norm}. 
We denote the entropy calculated from the output mask $B_{s_l}$ according to Section 3.4 of the main paper as $E_{s_l}$. In Tables~\ref{tab_network_arch_dnet} and~\ref{tab_network_arch_wnet}, we denote 2D convolutions as \textit{2D conv}, deformable 2D convolutions~\cite{deform_convsv2} as \textit{2D def. conv} and transposed 2D convolutions as \textit{2D tran. conv.} Further, we denote the leaky ReLU activation function as \textit{LReLU}, the number of input and output channels with $\#C_{\text{in}}$ and $\#C_{\text{out}}$, the stride with \textit{str.} and kernel size with $k$ (we add $\backslash b$ when no bias is used, the padding is set to $\frac{k-1}{2}$). 

\section{Supplementary qualitative results}

\begin{figure}[t]
\centering
\includegraphics[width=1.0\textwidth]{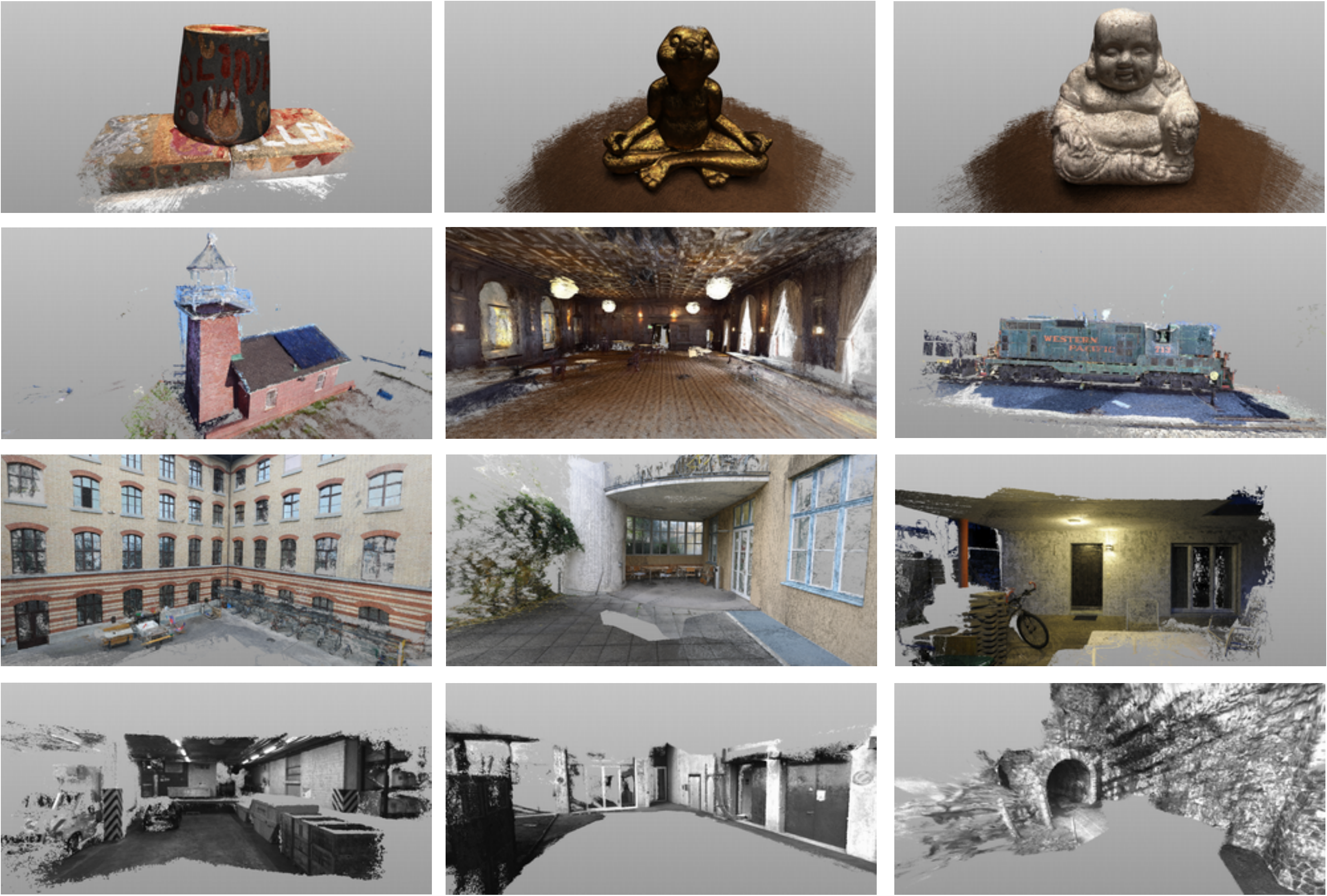}
\caption{Qualitative point cloud results for DTU~\cite{dtu} (first row), Tanks and Temples~\cite{tanksandtemples} (second row) and ETH3D~\cite{eth3d} high and low-res (third and fourth row).}
\label{fig_sup_point_cloud_vis}
\end{figure}

\begin{figure}[t]
\centering
\includegraphics[width=1.0\textwidth]{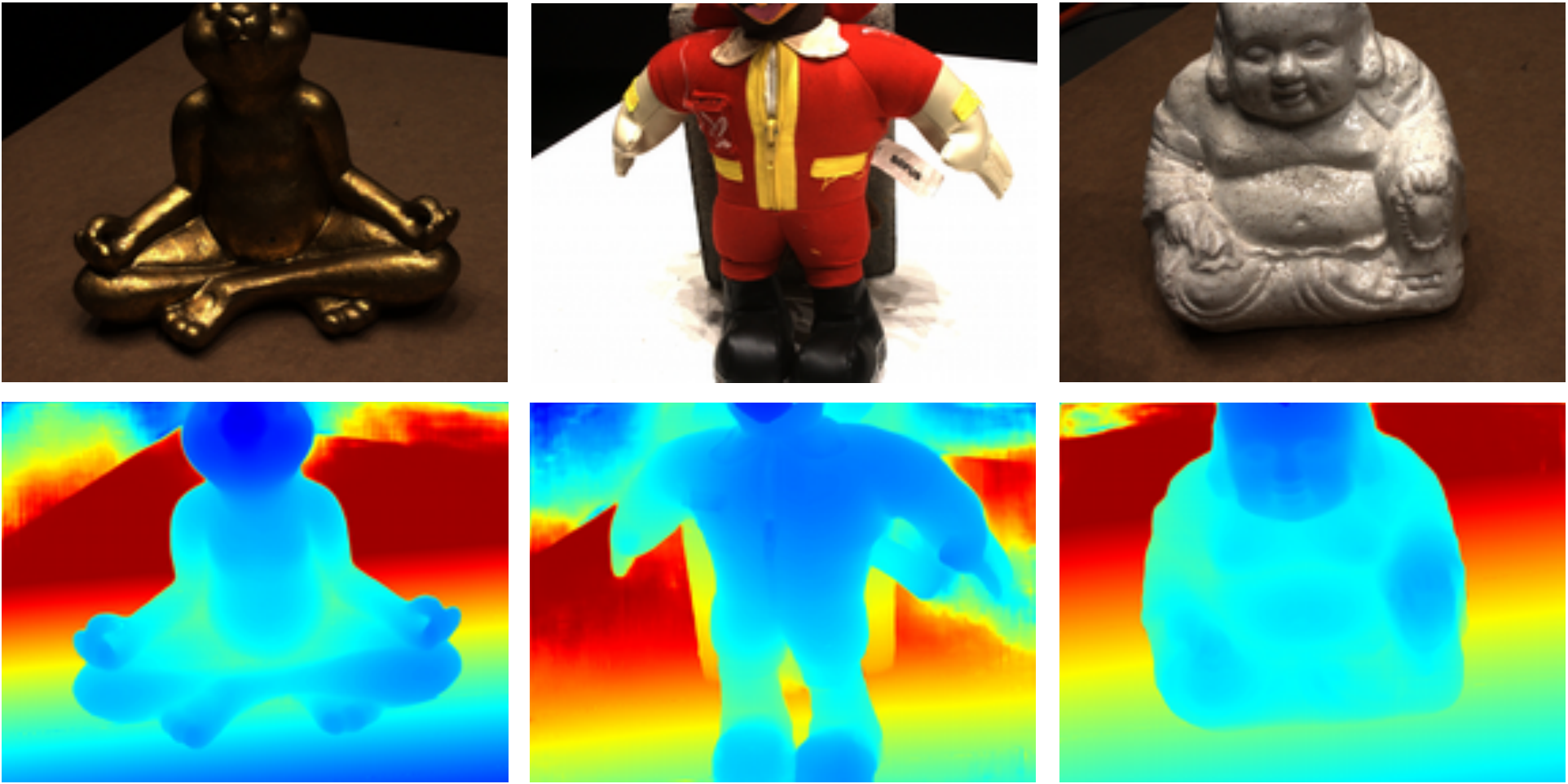}
\caption{Qualitative depth map results for DTU~\cite{dtu}. For each column, the reference image is at the top and the corresponding IB-MVS depth map at the bottom. Low to high depth values are color coded from blue to red.}
\label{fig_sup_depth_map_vis}
\end{figure}

In Figure~\ref{fig_sup_point_cloud_vis}, we provide qualitative point cloud results of our method IB-MVS for the DTU \cite{dtu}, Tanks and Temples \cite{tanksandtemples} and ETH3D~\cite{eth3d} high and low-res datasets. Additionally, in Figure~\ref{fig_sup_depth_map_vis}, we provide supplementary qualitative IB-MVS depth map results from the DTU~\cite{dtu} dataset. 

\begin{figure}[t]
\centering
\includegraphics[width=0.99\textwidth]{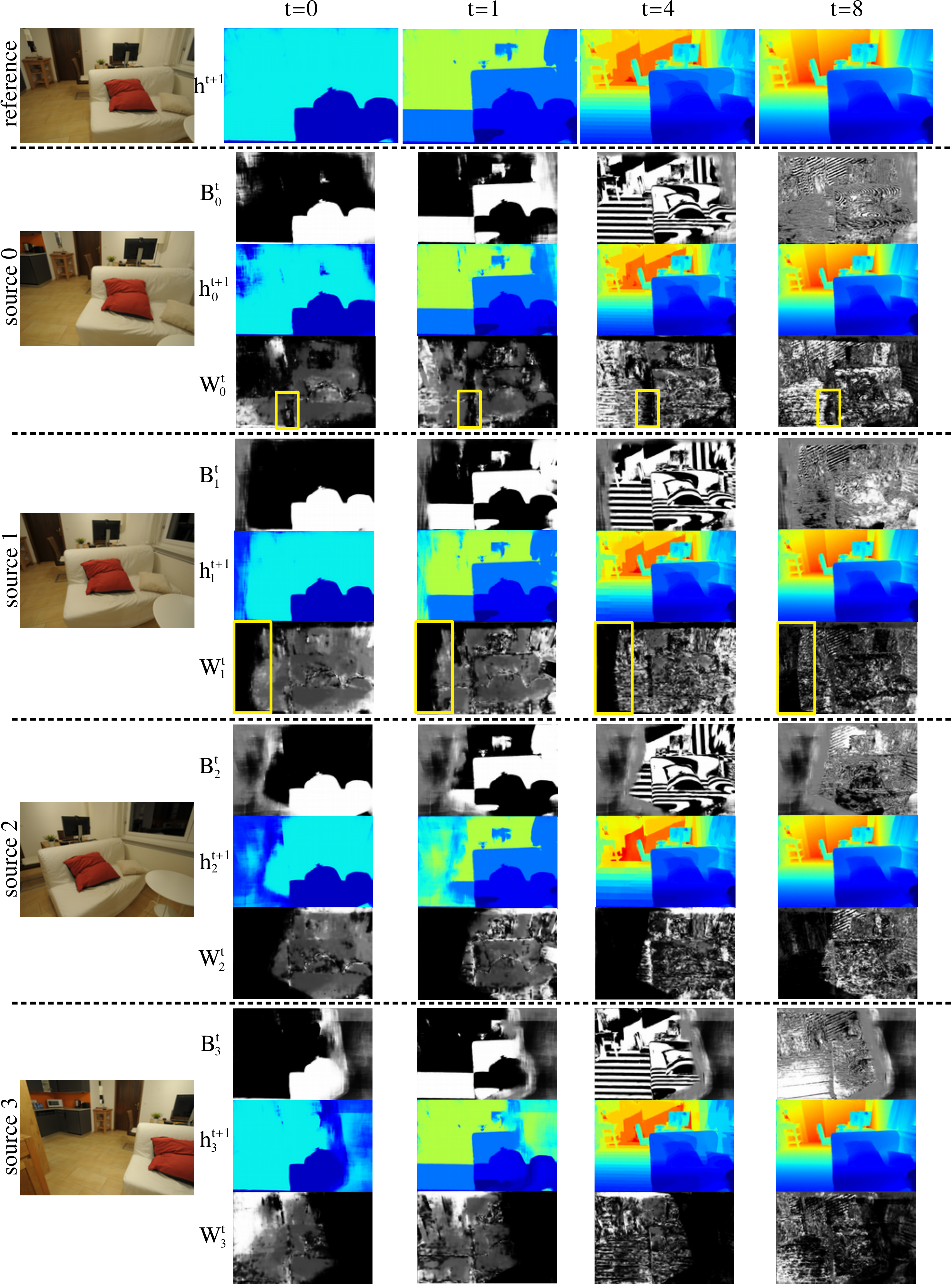}
\caption{We visualize intermediate results of IB-MVS for a view of the ETH3D~\cite{eth3d} high-res \texttt{living room} scene.
The top row shows the depth hypothesis $h^{t+1}$ predicted at iteration $t$ by fusing the depth hypothesis $h_s^{t+1}$ from different source images according to $W_s^t$. We also show the binary decision masks $B_s^t$. We color code low to high depth values from blue to red. For $B_s^t$ and $W_s^t$ black represents the value $0$ and white represents the value $1$. 
}
\label{fig_algo_results_sup}
\end{figure}

In order to provide further insights into IB-MVS, in Figure~\ref{fig_algo_results_sup} we provide a visualization of its intermediate outputs at different iterations $t = 0, 1, 4, 8$.
For each source image $I_s$, at iteration $t$ the hypothesis $h^t$ is used to compute the binary decision mask $B_s^t$ and the weight mask $W_s^t$ via D-Net and W-Net, respectively; this permits to compute the new reference depth map hypothesis $h_s^{t+1}$ using Eq. (2) of the main paper.
The rows 2-5 of Figure~\ref{fig_algo_results_sup} show $B_s^t$, $W_s^t$ and $h_s^{t+1}$ for the 4 source images and different values of $t$.
The new reference image depth hypothesis $h_s^{t+1}$ are then fused into a single depth map $h^{t+1}$ using the weights $W_s^t$.
The first row of Figure~\ref{fig_algo_results_sup} shows the reference image along with $h^{t+1}$ for different values of $t$.
We conclude by observing that, as desired, the weight masks in Figure~\ref{fig_algo_results_sup} assign a low confidence to those areas of the reference image that are occluded in the source image, as these areas cannot be matched.
This can be appreciated in the weight masks $W_s^t$ depicted in rows 2 and 3, where the area below the sofa and the left-most region (highlighted in yellow) of the reference image are dark because they are occluded in the respective source images.

\begin{table}[t] 
  \centering
  \small
  \setlength{\tabcolsep}{1pt}
  \begin{tabular}{@{}llccl@{}}
      \toprule
      in name & out name & $\#C_{\text{in}}$ & $\#C_{\text{out}}$ & operation \\ \midrule
      $\text{Feat}_{r_l}$ & Conv1 & $F_l$ & $F_l$ & 2D conv., k=3, str.=1, act.=LReLU \\
      $\text{Feat}_{s_l}$ & DConv1 & $F_l$ & $F_l$ & 2D def. conv., k=5, str.=1, act.=LReLU \\
      Conv1 | DConv1 & Conc1 & $2F_l$ & $2F_l$ & concatenate along channel dim. \\
      Conc1 & Conv2 & $2F_l$ & $2F_l$ & 2D conv., k=3, str.=1, act.=LReLU \\
      Conv2 & Sc1 & $2F_l$ & $2F_l$ & 2D conv., k=3, str.=2, act.=LReLU \\
      $\text{Feat}_{r_l}$ & $\text{Feat}_{r_l}$ half & $F_l$ & $F_l$ & bilinear interp. downscale to half \\
      $\text{Feat}_s{_l}$ & $\text{Feat}_{s_l}$ half & $F_l$ & $F_l$ & bilinear interp. downscale to half \\
      $\text{Feat}_{r_l}$ half & Conv3 & $F_l$ & $F_l$ & 2D conv., k=3, str.=1, act.=LReLU
      \\
      $\text{Feat}_{s_l}$ half & DConv2 & $F_l$ & $F_l$ & 2D def. conv., k=5, str.=1, act.=LReLU \\
      Conv3 | DConv2 & Conc2 & $2F_l$ & $2F_l$ & concatenate along channel dim. \\
      Conc2 & Conv4 & $2F_l$ & $2F_l$ & 2D conv., k=3, str.=1, act.=LReLU \\
      Sc1 | Conv4 & Conc3 & $4F_l$ & $4F_l$ & for $l=0$ concatenate along channel dim. \\
      Conc3 & Conv5 & $4F_l$ & $4F_l$ & for $l=0$ 2D conv., k=3, str.=1, act.=LReLU \\
      $\text{Fo}_{l-1}$ | Sc1 | Conv4 & Conc3 & $4F_l + 4F_{l-1}$ & $4F_l + 4F_{l-1}$ & for $l>0$ concatenate along channel dim. \\
      Conc3 & ConvPr & $4F_l + 4F_{l-1}$ & $4F_l$ & for $l>0$ 2D conv., k=3, str.=1, act.=LReLU \\
      ConvPr & Conv5 & $4F_l$ & $4F_l$ & for $l>0$ 2D conv., k=3, str.=1, act.=LReLU \\
      Conv5 & Sc2 & $4F_l$ & $4F_l$ & 2D conv., k=3, str.=2, act.=LReLU \\
      $\text{Feat}_{r_l}$ & $\text{Feat}_{r_l}$ quar. & $F_l$ & $F_l$ & bilinear interp. downscale to quarter \\
      $\text{Feat}_s{_l}$ & $\text{Feat}_{s_l}$ quar. & $F_l$ & $F_l$ & bilinear interp. downscale to quarter \\
      $\text{Feat}_{r_l}$ quar. & Conv6 & $F_l$ & $F_l$ & 2D conv., k=3, str.=1, act.=LReLU
      \\
      $\text{Feat}_{s_l}$ quar. & DConv3 & $F_l$ & $F_l$ & 2D def. conv., k=5, str.=1, act.=LReLU \\
      Conv6 | DConv3 & Conc4 & $2F_l$ & $2F_l$ & concatenate along channel dim. \\
      Conc4 & Conv7 & $2F_l$ & $2F_l$ & 2D conv., k=3, str.=1, act.=LReLU \\
      Sc2 | Conv7 & Conc5 & $6F_l$ & $6F_l$ & concatenate along channel dim. \\
      Conc5 & Conv8 & $6F_l$ & $6F_l$ & 2D conv., k=3, str.=1, act.=LReLU \\
      Conv8 & Conv9 & $6F_l$ & $6F_l$ & 2D conv., k=3, str.=1, act.=LReLU \\
      Conv9 & Conv10 & $6F_l$ & $6F_l$ & 2D conv., k=3, str.=1, act.=LReLU \\
      Conv10 & UConv1 & $6F_l$ & $6F_l$ & 2D tran. conv.,k=4$\backslash b$, str.=2, act.=LReLU  \\
      Conv5 | UConv1 & Conc6 & $10F_l$ & $10F_l$ & concatenate along channel dim. \\
      Conc6 & Conv11 & $10F_l$ & $4F_l$ & 2D conv., k=3, str.=1, act.=LReLU \\
      Conv11 & Conv12 & $4F_l$ & $4F_l$ & 2D conv., k=3, str.=1, act.=LReLU \\
      Conv12 & UConv2 & $4F_l$ & $4F_l$ & 2D tran. conv.,k=4$\backslash b$, str.=2, act.=LReLU  \\
      Conv2 | UConv2 & Conc7 & $6F_l$ & $6F_l$ & concatenate along channel dim. \\
      Conc7 & $\text{Fo}_{l}$ & $6F_l$ & $4F_l$ & 2D conv., k=3, str.=1, act.=LReLU \\
      $\text{Fo}_{l}$ & $B_{s_l}$ & $4F_l$ & $1$ & 2D conv., k=3$\backslash b$, str.=1, act.=sigmoid \\
      \bottomrule
      \end{tabular}
  \caption{D-Net architecture hyper-parameters, specifying the convolution type, number of input and output channels, kernel size, stride and activation function.}
  \label{tab_network_arch_dnet}
  \end{table}

\FloatBarrier

 \begin{table}[t] 
  \centering
  \small
  \setlength{\tabcolsep}{2pt}
  \begin{tabular}{@{}llccl@{}}
      \toprule
      in name & out name & $\#C_{\text{in}}$ & $\#C_{\text{out}}$ & operation \\ \midrule
      $E_{s_l}$ & Conv1 & $1$ & $2F_l$ & for $l=0$ 2D conv., k=3, str.=1, act.=LReLU \\
      $E_{s_l}$ & Conv0 & $1$ & $F_l$ & for $l>0$ 2D conv., k=3, str.=1, act.=LReLU \\
      $\text{Fo}_{l-1}$ & $\text{Fo}_{\text{Up}}$ & $\frac{F_{l-1}}{2}$ & $\frac{F_{l-1}}{2}$ & for $l>0$ bilinear interp. upscale to double res. \\
      $\text{Fo}_{\text{Up}}$ & ConvPr & $\frac{F_{l-1}}{2}$ & $F_l$ & for $l>0$ 2D conv., k=3, str.=1, act.=LReLU \\
      Conv0 | ConvPr & Conc1 & $2F_l$ & $2F_l$ & for $l>0$ concatenate along channel dim. \\
      Conc1 & Conv1 & $2F_l$ & $2F_l$ & for $l>0$ 2D conv., k=3, str.=1, act.=LReLU \\
      Conv1 & Conv2 & $2F_l$ & $2F_l$ & 2D conv., k=3, str.=1, act.=LReLU \\
      Conv2 & Conv3 & $2F_l$ & $F_l$ & 2D conv., k=3, str.=1, act.=LReLU \\
      Conv3 & $\text{Fo}_{l}$ & $2F_l$ & $\frac{F_l}{2}$ & 2D conv., k=3, str.=1, act.=LReLU \\
      $\text{Fo}_{l}$ & $w_{s_l}$ & $\frac{F_l}{2}$ & 1 & 2D conv., k=3$\backslash b$, str.=1, act.=identity \\
      \bottomrule
      \end{tabular}
  \caption{W-Net architecture hyper-parameters, specifying the convolution type, number of input and output channels, kernel size, stride and activation function.}
  \label{tab_network_arch_wnet}
  \end{table}

\FloatBarrier

\bibliography{references}
\end{document}